\documentclass[letterpaper, 10 pt, conference]{ieeeconf}  

\IEEEoverridecommandlockouts                              

\usepackage{graphics} 
\usepackage{epsfig} 
\usepackage{times} 
\usepackage{amsmath} 
\usepackage{amssymb}  
\usepackage{bbding}
\usepackage{booktabs}
\usepackage{algorithm}
\usepackage{algpseudocode}
\usepackage{pifont}
\usepackage{color}
\newcommand{\xmark}{\ding{55}}
\usepackage{tabularx}
\usepackage{booktabs}

\title{\LARGE \bf
Rearranging the Environment to Maximize Energy with a Robotic Circuit Drawing
}

\author{Xianglong Tan$^{1,2,*}$, Zhikang Liu$^{1,3,*}$, Chen Yu$^{1}$ and Andre Rosendo$^{1}$
\thanks{*This work was supported by the NNSFC grant 61850410527.}
\thanks{$^{1}$Living Machines Laboratory, School of Information Science and Technology, ShanghaiTech University, China.
        }%
\thanks{$^{2}$Hamlyn Centre,
        Imperial College London, UK.
        }%
\thanks{$^{3}$Department of Computer Science,
        University of York, UK.
        }%
\thanks{$^{*}$These authors contributed equally to this work.
        }%
}

\begin{document}

\maketitle
\thispagestyle{empty}
\pagestyle{empty}

\begin{abstract}
Robots with the ability to actively acquire power from surroundings will be greatly beneficial for long-term autonomy and to survive in uncertain environments. In this work, we present a robot capable of drawing circuits with conductive ink while also rearranging the visual world to receive maximum energy from a power source. A range of circuit drawing tasks is designed to simulate real-world scenarios, including avoiding physical obstacles and regions that would discontinue drawn circuits. We adopt the state-of-the-art Transporter networks for pick-and-place manipulation from visual observation. We conduct experiments in both simulation and real-world settings, and our results show that, with a small number of demonstrations, the robot learns to rearrange the placement of objects (removing obstacles and bridging areas unsuitable for drawing) and to connect a power source with a minimum amount of conductive ink. As autonomous robots become more present, in our houses and other planets, our proposed method brings a novel way for machines to keep themselves functional by rearranging their surroundings to create their own electric circuits.

\end{abstract}

\section{INTRODUCTION}

Recent developments in the fields of robotic hardware, sensing, and machine learning have led to great progress in robotic applications in various areas \cite{burger2020mobile, duckett2018agricultural,9310688}. Recent advances in Robotics have led to progress in various areas and naturally raised the demand for fully-autonomous, long-lasting and self-evolving robots \cite{10.1371/journal.pone.0186107}, keeping human efforts further out of the loop. As with living species, robots consume energy to perform different tasks. Industrial robots deployed in the real world are typically powered by an uninterruptible power supply without the danger of running out of energy. For mobile robots, however, have their mobility, operational time and performance limited by the low storage capacity of batteries \cite{Yangeaar7650}. This prompts the benefits of an adaptable energy acquisition ability for robots operating in inhospitable environments or away from human aid.

Previous works on self-recharging robots showed a framework for robots to navigate autonomously to a charging station \cite{4413695}. They use cyclic genetic algorithm to optimize the control program in simulation. In another work \cite{5509556}, researchers built an autonomous robotic system capable of plugging itself into electrical outlets to recharge. Mayton et al. \cite{5509643} proposed a mobile manipulation platform capable of plugging itself into a standard U.S. electrical outlet. Instead of using vision to assist the process, the plugging was guided entirely by measurements of electrical emissions from the outlets. However, outlets have different standards across regions and they are not always available, especially in the field and unexplored areas. Batteries are more often used in these scenarios, but the connection of cables to batteries by robots is difficult and a fairly unexplored field of research. Other researchers focus on the design of power systems for self-powered robots. Solar panels are widely used for robots to harvest energy from ambient sources to recharge batteries. They are normally low-cost and light-weight, but suffer from low efficiency and restrictive application environment. In a series of works on energy harvesting \cite{Pan1947, Wang102, Xu2010}, researchers demonstrate the powering of nanobots from ambient mechanical energy. In a more recent work \cite{doi:10.1021/acsenergylett.9b02661}, a novel robotic power system is powered by scavenging energy from external metals. However, in these works the power harvested from the environment is very limited, and it can be hardly used to power any standard robot.

Manipulation of objects and rearrangement of environments have a long history in robotics. Traditional pick-and-place solutions follow a standard pipeline: object recognition, pose estimation, and then grasp planning. For instance, Jonschkowski et al. \cite{jonschkowski2016probabilistic} used handcrafted image features for object perception followed by computation of pose estimation for suction objects. Recent data-driven solutions for manipulation tasks involve imitation learning (IL) and reinforcement learning (RL). Finn et al. \cite{finn2017one} used meta-learning as a pre-training procedure that uses demonstrations from different environments. The method is shown to be able to achieve one-shot imitation learning for novel scenarios and scale to raw pixel inputs. Duan et al. \cite{NIPS2017_ba386660} also achieve one-shot imitation on a block stacking task. They used soft attention for processing both the state-action sequence that corresponds to the demonstration, and for processing the components of the vector specifying the locations of various blocks. RL algorithms have also shown their ability on learning reliable policy end-to-end with raw image inputs. Kalashnikov et al. \cite{Kalashnikov2018} proposed an RL algorithm that is entirely self-supervised, using only grasp outcome labels that are obtained automatically by the robot for the grasping task. Zeng et al. \cite{zeng2018learning} performed RL to learn complementary pushing and grasping policies that operate end-to-end from visual observations to actions. Besides using IL or RL,  Zeng et al. \cite{zeng2020transporter} also proposed a framework called Transporter Networks that rearranges deep features to infer spatial displacements from visual input, which can parameterize robot actions. Experimental results show that it is orders of magnitude more sample efficient than a variety of end-to-end baselines. 

This paper presents a novel approach for robots to leverage ambient electrical power sources and survive in resource-limited, uncertain environments. Instead of using wires or cables, the robot learns to rearrange environmental obstacles and useful resources to build optimal electrical path to a power supply. We develop a path planning algorithm for 3D circuit drawing and adopt a novel network, capable of selecting pick and place pose for the robot to reposition objects. This paper shows that robots can rearrange the physical world to receive energy from the environment to stay alive. 

\section{ROBOTIC CIRCUIT DRAWING}

\subsection{Conductive Ink}

Graphene-based conductive ink has shown great potential in printing flexible electronics for its low cost, high connectivity and can be applied directly on materials like textile\cite{C7TC03669H}, paper, and other diverse flexible substrates\cite{Huang:2011aa}. Compared to metal-based conductive ink, it is low-toxic, environment friendly, and easy to make and store \cite{Pan_2018}. In this work, we follow the instructions in \cite{Saidina_2019} to fabricate customized graphene-based conductive ink. The ink is made of $5$ $wt\%$ graphene flakes, $0.5$ $wt\%$ graphene dispersion, and $94.5$ $wt\%$ water, which has a sheet resistance of approximately $2$ $\Omega/sq$.

\subsection{Simulation Setup and Tasks}
The simulation is set up with PyBullet \cite{pybullet}. We use a Kinova 6 DoF Jaco robot arm to perform circuit drawing on a 0.8$\times$0.6m workspace. The two-finger gripper of the robot enables picking and placing of objects. The electrodes of the robot and the power source are placed on opposite sides of the workspace. The main goal is to draw continuous ink to connect the electrode pair of the robot and the power source. 

\begin{figure*}[t!]

\centerline{\includegraphics[width=1\textwidth]{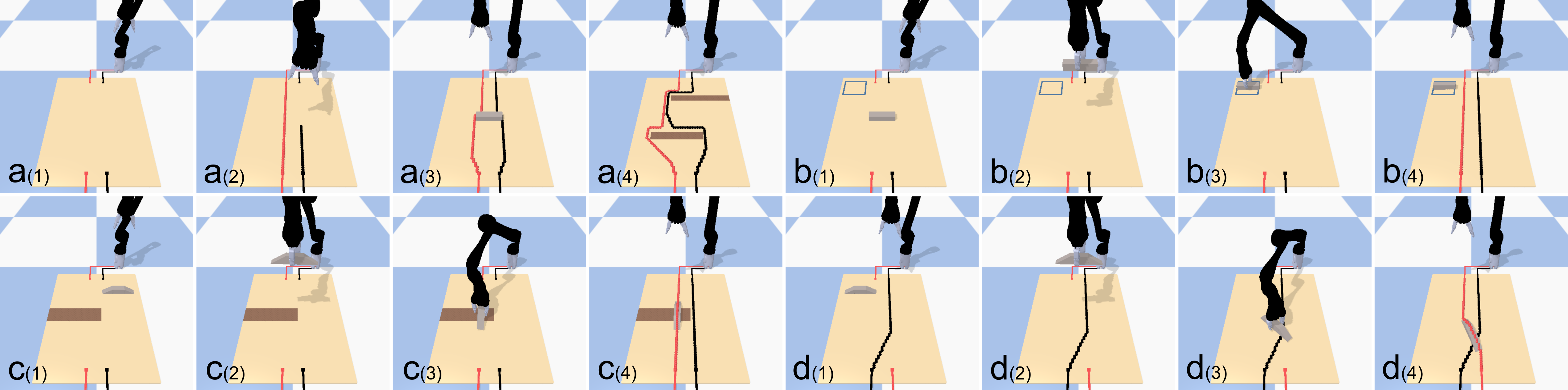}}
\caption{The four basic circuit drawing tasks in this paper: path planning for circuit drawing (a), pick and remove obstacles (b),  reposition ramps to draw through a forbidden zone (brown area) (c), reposition ramps to draw above circuits (d). The workspace is a 0.8$\times$0.6m platform with four electrodes (two from the robot arm and two from a power source) on each side (a$_1$). The basic goal for the robot is to connect electrodes with conductive ink (a$_2$), while avoiding obstacles (a$_3$) and forbidden zones (a$_4$). In task (b), the robot should pick the obstacle (b$_2$) and place it in a zone away from the center (b$_3$) to draw the shortest circuit between electrodes (b$_4$). In task (c), when a forbidden zone is presented (c$_1$), the robot is required to pick a ramp (c$_2$) and place it in the zone (c$_3$) to allow drawing circuit through the zone (c$_4$). In task (d), the electrodes of the robot and the power source are in the opposite position, the robot needs to connect a pair of electrode first (d$_1$), then pick a ramp (d$_2$) and place on the drawn circuit (d$_3$) to connect the rest electrodes without shortcircuiting the power source (d$_4$)}
\label{fig:tasks}
\end{figure*}

Circuit drawing tasks are selected based on the possible situation a robot confronts in complex real-world environments. For example, in search and rescue missions, the robot needs to automatically plan the drawing path avoiding obstacles and areas that would discontinue or damage the circuit (e.g., huge bumps, holes, surface with liquid, sand, etc). We refer to these areas as forbidden zones in this paper. Meanwhile, drawing longer the circuit path leads to higher circuit resistance and more ink and energy used. If any blocking obstacles can be moved by the robot, they should be removed before the drawing. Moreover, if movable bridge-like structures are presented in the workspace, reposition them on top of any forbidden zones could enable drawing above these zones to connect more efficiently. The details of circuit drawing tasks are shown in Fig.~\ref{fig:tasks}.

\section{METHODS}



\subsection{Transporter Networks}

We consider the pick-and-place problem with actions $\mathbf{a}_t$ defining the pick and place poses and states $\mathbf{s}_t$ represented by visual observations:
\begin{equation}
\pi\left(\mathbf{s}_{t}\right) \rightarrow \mathbf{a}_{t}=\left(\mathcal{T}_{\text {pick}}, \mathcal{T}_{\text {place }}\right) \in \mathcal{A},
\end{equation}
where $\mathcal{T}_{\text{pick}}$ is the pose of the end effector used to pick an object, and $\mathcal{T}_{\text {place}}$ to place the object, where $\mathcal{T}_{\text{pick}}, \mathcal{T}_{\text {place }} \in \mathbb{R}^{2}$ are 2D coordinates. In our task, the goal is to pick up the bridge-like blocks then place them over the forbidden zones while picking up other blocks then placing them out of the working area.

Other than using end-to-end RL for solving the 
pick-and-place task
\cite{zhang2020grasp, singh2019, nair2018visual, zeng2018learning}, we use Transporter Networks \cite{zeng2020transporter} to recover the distribution of successful pick poses
and the corresponding distribution of successful place poses 
. The distributions are trained with pure visual observations, which allows no assumption of prior information about the picking objects. The state $\mathbf{s}_t$ in our experiment is a grid of pixels reconstructed from RGB-D information. We use the pick-and-place actions in the space of 2D translations. We discretize the space of SO(2) rotations into $k$ bins where the input visual observation
$\mathbf{s}_t$ are rotated. $\mathbf{s}_t$ is therefore defined on a grid of $\{(x,y,w)_i\}$ where $w$ describes the rotation.

Denote such a pose $\{(x,y,w)i\}$ as $\xi$. We use Q value function $\mathcal{Q}_{\mathrm{pick}}(\xi|\mathbf{s}_t)$ to model the distribution of the returns among different picking positions and $\mathcal{Q}_{\mathrm{place}}(\xi|\mathbf{s}_t,\mathcal{T}_{\text{pick}})$ for placing positions.

\textbf{Learning Picking and Placing}
After the Q function $\mathcal{Q}_{\mathrm{pick}}(\xi|\mathbf{s}_t)$ is trained by a fully convolutional network (FCN), the picking position can be obtained by
\begin{equation}
\mathcal{T}_{\text {pick }}=\underset{\xi}{\operatorname{argmax}} \mathcal{Q}_{\text {pick }}\left(\xi \mid \mathbf{s}_{t}\right).
\end{equation}
Denote $\mathbf{s}_{t}[\xi]$ as a partial crop centered on a pose $\xi$. We search for the best position for placing that match with the highest feature correlation. The Q function for placing can then be written as
\begin{equation}
\mathcal{Q}_{\text {place }}\left(\xi \mid \mathbf{s}_{t}, \mathcal{T}_{\text {pick }}\right)=\psi\left(\mathbf{s}_{t}\left[\mathcal{T}_{\text {pick }}\right]\right) * \phi\left(\mathbf{s}_{t}\right)[\xi]
\label{eq:cross}
\end{equation}
where $\psi(\cdot)$ and $\phi(\cdot)$ are dense feature embeddings from two deep models. Similarly, the pose of placing is hence obtained by 
\begin{equation}
\mathcal{T}_{\text {place }}=\underset{\xi}{\operatorname{argmax}} \mathcal{Q}_{\text {place }}\left(\xi \mid \mathbf{s}_{t}, \mathcal{T}_{\text {pick }}\right)
\end{equation}

\textbf{Network Architecture} We use visual observation as the state $\mathbf{s}_t$, which is a $320\times240$ RGB-D image
, where each pixel contains both color and distance information. This enables our models to learn both texture features and shape features.
The picking model is a FCN that takes the state $\mathbf{s}_{t} \in \mathbb{R}^{H \times W \times 4}$ as input and outputs dense pixel-wise state value  $\mathcal{V}_{\text {pick }} \in \mathbb{R}^{H \times W}=\mathrm{softmax}(\mathcal{Q}_{pick}(\xi|\mathbf{s}_t))$. Specifically, the picking model is an hourglass network with encoder-decoder architecture. It is a 43-layer residual network \cite{he2016deep} with 12 residual blocks, 8-stride, and image-wide softmax. Dilation \cite{Yu2016} and ReLU activations \cite{Nair2020} are implemented on all layers except the first. It acts as an attention mechanism. 
The placing model is a two-stream FCN that takes the state $\mathbf{s}_{t} \in \mathbb{R}^{H \times W \times 4}$ as input and outputs two dense feature maps:  features$\psi\left(\mathbf{s}_{t}\right) \in \mathbb{R}^{H \times W \times d}$ and  $\phi\left(\mathbf{s}_{t}\right) \in \mathbb{R}^{H \times W \times d}$ , where $d$ is the number of feature dimensions. The architecture of the placing model is similar to the picking one but without ReLU activations in the last layer. 
Once the picking pose is obtained by $\mathcal{T}_{\text {pick }}=\operatorname{argmax} \mathcal{V}_{\text {pick }}$, the partial crop with size $c$,  $\psi\left(\mathbf{s}_{t}\left[\mathcal{T}_{\text {pick }}\right]\right) \in \mathbb{R}^{c \times c \times d}$ centered around $\mathcal{T}_{\text {pick }}$ is transformed by $\xi$ and cross-correlated with the feature map $\phi\left(\mathbf{s}_{t}\right)$ according to (\ref{eq:cross}). The pixel-wise placing state value can then be obtained by $\mathcal{V}_{\text {place }} \in \mathbb{R}^{H \times W \times k}=\operatorname{softmax}\left(\mathcal{Q}_{\text {place }}\left(\xi \mid \mathbf{s}_{t}, \mathcal{T}_{\text {pick }}\right)\right)$. $\mathcal{T}_{\text {place }}=\operatorname{argmax} \mathcal{V}_{\text {place }}$ is the final placing pose that has the highest correlation with two feature partial crops.

\textbf{Learning from Demonstrations}
We use the rewards $\mathcal{R}$ to collect data for a dataset $\mathcal{D}$ of $n$ expert-demonstrated trajectories $\tau$, where each trajectory $\tau_{i}$ 
is a sequence of state-action pairs $(\mathbf{s}_t,\mathbf{a}_t)$. During the training process, we
sample state-action pairs from the dataset $D$ serving as $\mathcal{T}_{\text{pick}}$ and $\mathcal{T}_{\text{pick}}$ labels. The one-hot pixel maps $Y_{\text {pick }} \in \mathbb{R}^{H \times W}$ and $Y_{\text {place }} \in \mathbb{R}^{H \times W \times k}$ the are generated by two training labels. The loss function is the cross-entropy between these one-hot pixel maps and the state values of picking and placing: 
\begin{equation}
\mathcal{L}=-\mathbb{E}_{Y_{\text {pick }}}\left[\log \mathcal{V}_{\text {pick }}\right]-\mathbb{E}_{Y_{\text {place }}}\left[\log \mathcal{V}_{\text {place }}\right].
\end{equation}
We used 3D path planning for calculating the reward ${r_t \in \mathcal{R}}$.

\subsection{3D Path Planning}

The path planning is used to not only make the shortest path of the drawing circuit on the ground, but also be treated as a filter to select the high-quality data samples for training.

\textbf{Information Input} The terrain information will just be a visual observation, which is an orthographic top-down view from the agent, and input as a 2D array, representing the situation of a 3-dimensional working area. It is different from 2-dimensional space, there is an additional consideration of displacement in the z index that should be added when we process the algorithm.

\textbf{Algorithm} In this case, An double layer A* search algorithm, which is an informed searching algorithm, has been used. Firstly, we start the A* search from the departure point, but we don't expand the points to the nearest points. It expands to the available grooves instead, whatever it is uphill or downhill, or choose directly to the destination if it is possible and in the same level. Then, the A* will calculate the distance between two nodes generated in the previous layer in the second layer.
\begin{figure}[ht]
\centering
\includegraphics[scale=0.25]{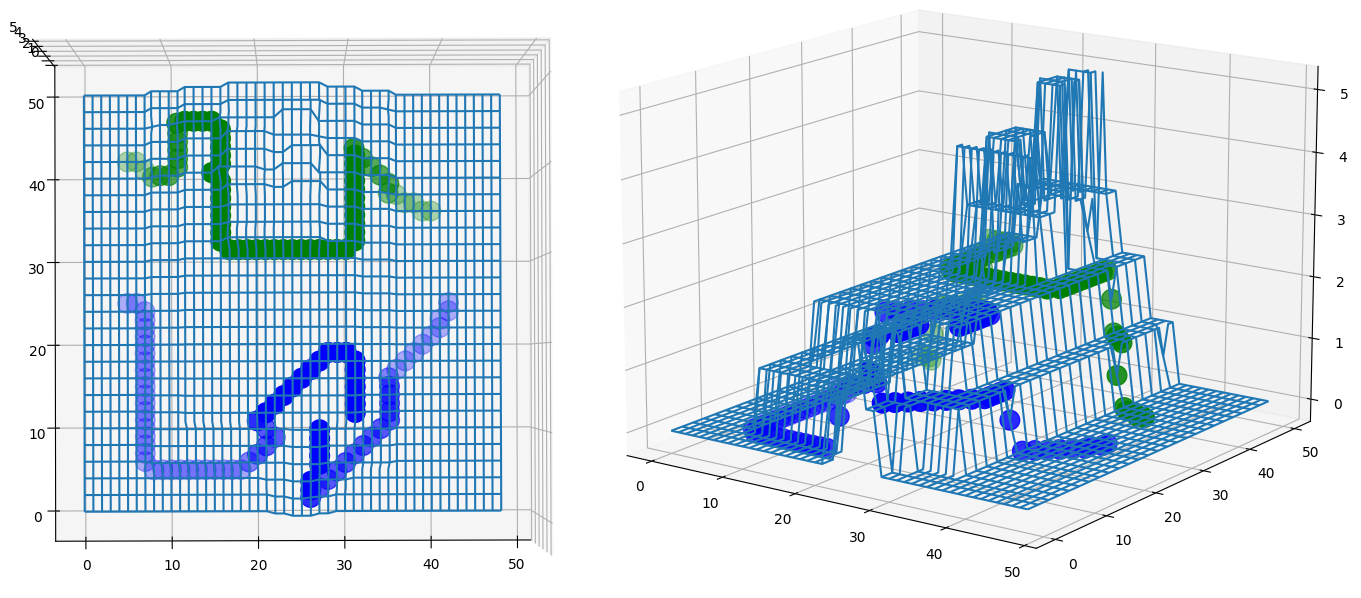}
\caption{Our 3D path planning algorithm offers the paths produced several features: 1. The path can go through 3D dimensional space 2. The path can only ascend or descend if the height difference of the coming node is lower than a predefined value. This fits into the situation when we are drawing a circuit line. 3. The algorithm is based on a double-layer A* algorithm, in the first layer, it expands to the available grooves, or chooses directly to the destination if it is possible and in the same level. In the second layer, it uses A* to calculate the displacement path from each node to the next node from the previous layer.}
\label{fig:pathplanning}
\end{figure}

This algorithm fits our path planning, which also serves for drawing circuit lines on the ground. The path is also not allowed to go through any steep edges, including cliffs and precipitous slopes. This is solved by defining a minimum and a maximum value to detect slope values. If the depth difference of two adjacent nodes is less than the minimum slope value, then we can treat it as they are at the same level. If it is greater than the minimum but less than the maximum, then it is treated as a slope to move on. And otherwise, it can not move on at all.

\begin{equation}
f_{\left(n\right)} = g_{\left(n\right)} + h_{\left(n\right)}
\end{equation}

Here is the equation when we use A* algorithm and calculate the cost of traveling to a node, where n is the node, g(n) is the cost from the parent node, and h(n), which is known as the Heuristic equation, is the predicted cost to the final destination. In a 2d path planning problem, the eulerian equation is just equal to $\sqrt{x^2 + y^2}$, but it is going to be $\sqrt{x^2 + y^2 + z^2}$ when it is in 3d space.


\subsection{Simulation and Training Details}

We use PyBullet \cite{pybullet} as the simulator and create a customized OpenAi gym \cite{gym} environment for all the tasks. We adopt the above-mentioned transporter networks as the learning agent. Our visual observation o$_t$ is an orthographic top-down view of a 0.8$\times$0.6 workspace, generated in PyBullet by fusing 640$\times$480 RGB-D images captured with calibrated cameras using known intrinsics and extrinsics. The top-down image o$_t$ has a pixel resolution of 320$\times$240 - each pixel represents a 2.5$\times$2.5mm vertical column of 3D space in the workspace. The number of rotation for kernels in transporter networks is set to 8. For all experiments, the initial position, color and type of objects are randomly set. The output action a$_t$ is the pick pose and the place pose of the gripper.

Samples for training are automatic collected and filtered only if the reward feedback of the action meet a threshold. The reward calculation have used the concept of capacitor discharging equation.

\begin{equation}
reward_{one circuit} = {e}^{\frac{mincost - cost}{cost} \times \mu} \\
\end{equation}
\begin{equation}
total_reward = reward_{A} \times weight_{A} + reward_{B} \times weight_{B}
\end{equation}

We train the networks for 1k iterations with 1, 10, and 100 demonstrations respectively and test the performance with 50 test samples for all the tasks. In each test, the number of steps for pick and place action equals to the movable objects in the workspace. If the position of electrodes are inverted, the robot will connect one of the electrode pairs one step before the maximum action steps. In other cases, the robot only start drawing circuits after all action steps.

\subsection{Sim-to-Real Experiment Setup}

\begin{figure}[t!]
\centerline{\includegraphics[width=0.48\textwidth]{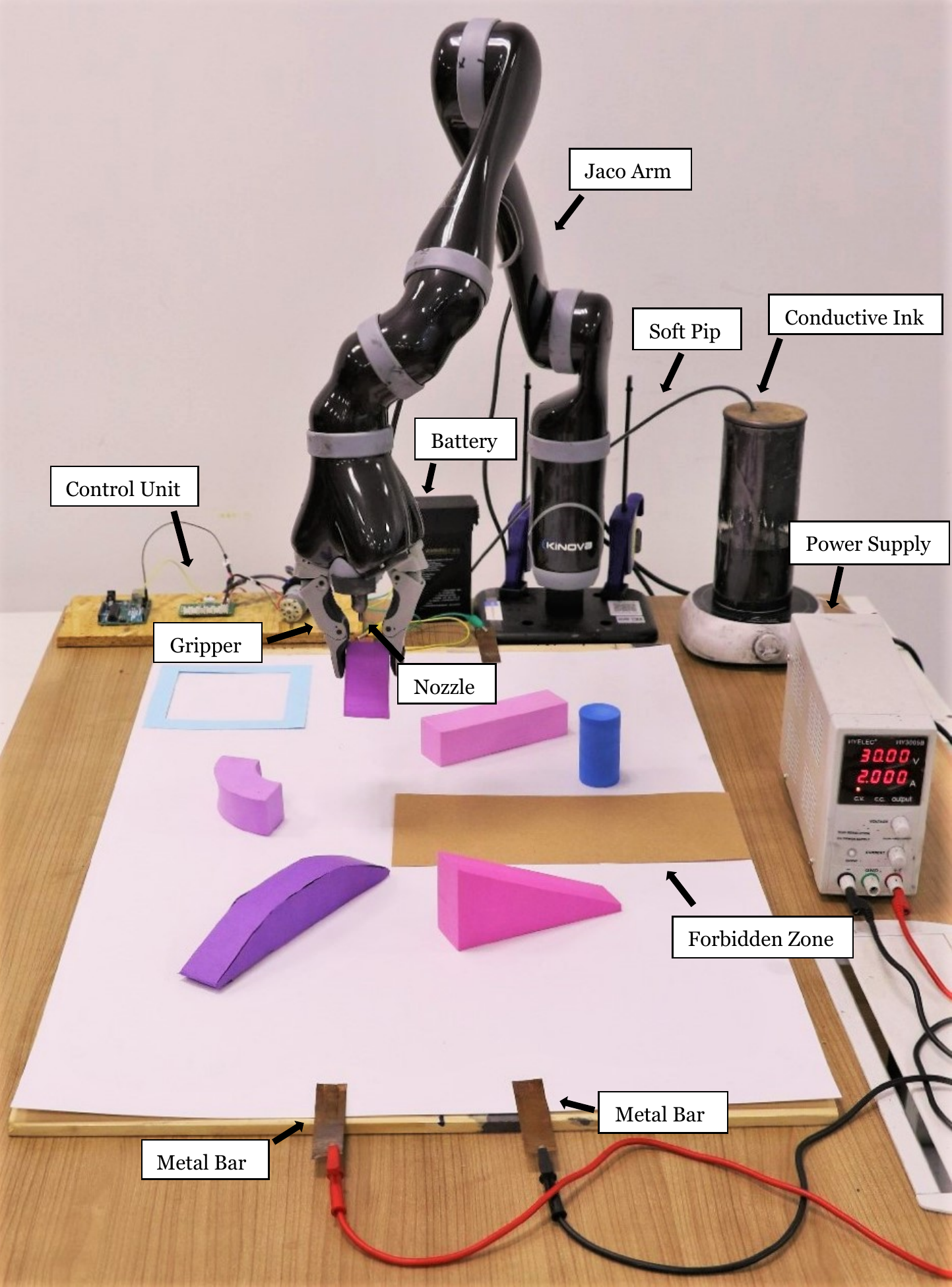}}
\caption{Experimental setup of the circuit drawing robot. The Kinova 6DOF Jaco Arm first performs pick-and-place actions at the pose suggested by the network. Then the arm moves to one of the metal bars at 5cm above the cardboard. The ROS controller continuously sends the waypoints of the circuit path calculated from the 3D path planning algorithm. An Arduino then receives the state of the arm through ROS ad sets the on/off of the peristaltic pump to control the ink flow. The connection starts to show conductivity after the ink dries (30 minutes).}
\label{fig:real_setup}
\end{figure}

The experimental setup is shown in Fig.~\ref{fig:real_setup}. The workspace is a 0.8$\times$0.6m wooden platform. The cardboard is placed on top of the platform and two metal bars are fixed on each side of the cardboard, representing the terminals of the robot and the power source. The conductive ink is prepared and stored in a glass jar placed on a magnetic stirrer to prevent the graphene from solidification. The glass jar is linked to a soft pipe that is connected to a peristaltic pump. The pump pushes the ink towards a nozzle which is held by a 3D-printed dispenser at the end-effector of the Kinova 6DOF Jaco Arm. An Arduino Uno is used to control the speed of ink flow. The robot arm is controlled by Moveit (https://github.com/ros-planning/moveit). We use one of the successful cases from the all-in-one task, which has two obstacles, one forbidden zone, and inverted electrodes, as the demonstration of the real-world performance. To achieve sim-to-real control, we use the objects with the same size and shape trained in the simulation to reconstruct the real-world workspace. Each pick and place pose is given by the trained networks. We used the built-in inverse kinematics solver of Moveit to move the robot to the desire poses.

\section{RESULTS}

\subsection{Simulation Tasks}
In the simulation environment, we have created and contributed a wide variety of tasks, which is shown in \ref{simulatiom_result_table}. As the difficulty of the task increases, the number of obstacles and forbidden zones has increased as well. We have trained the network in these simulations for 1000 iterations, with respectively 1, 10, and 100 demos.

\textbf{Trained models}  There are two types of models trained from the network, including including the attention model and the transport model, are responsible for picking and placing actions respectively. As the observation information inputs, the model will automatically find the maximum Q value and output the pose, including the position and orientation. Then, these models are able to act.

\begin{figure}[ht]
\centering
\includegraphics[scale=0.3]{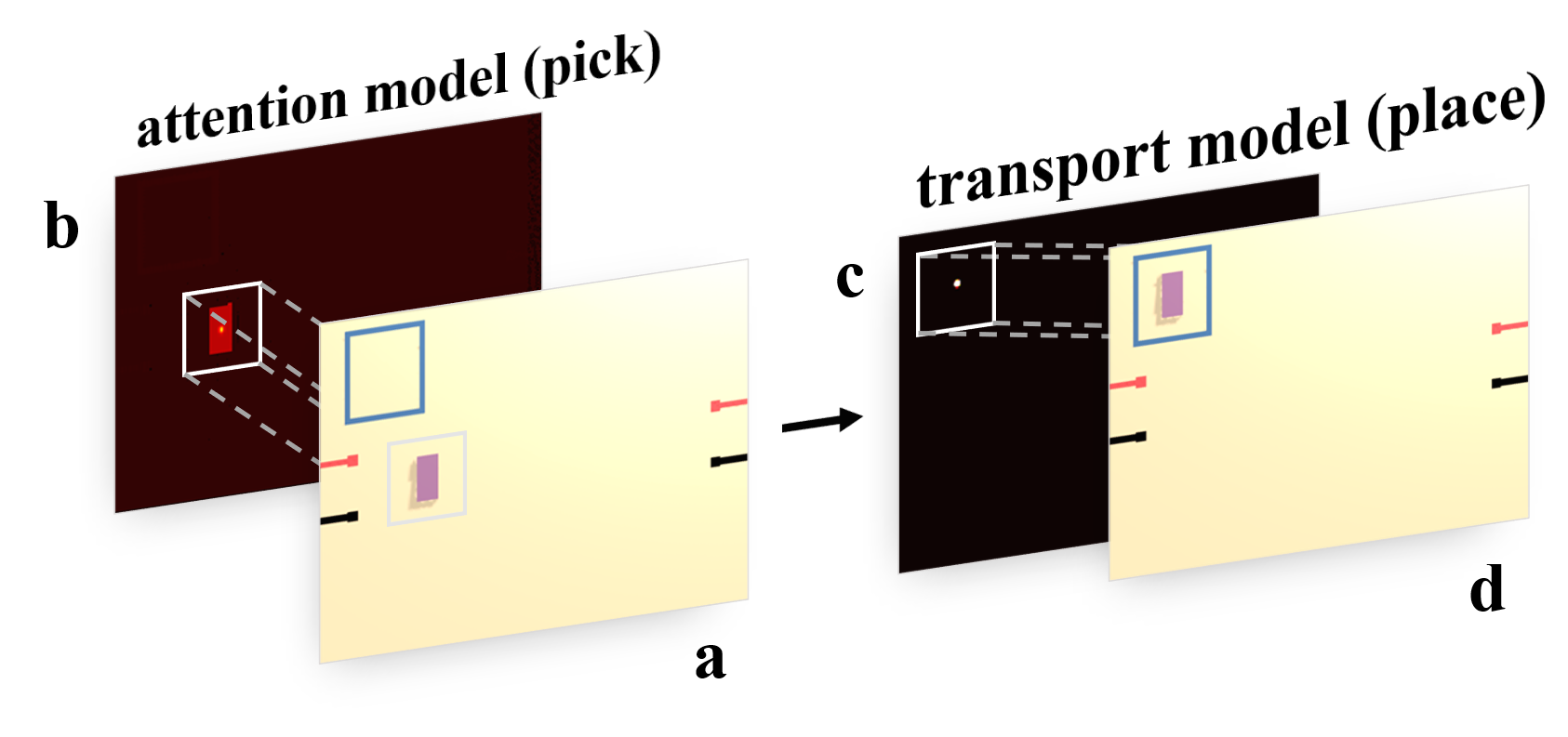}
\caption{In this situation (a) where the task is to pick up an obstacle, meanwhile try to shorten the length of the circuit path that is going to be drawn, and place the obstacle to the location we have set up on the top-left corner. The visual observation of this situation will be input and trained two models, including the attention model (b) and the transport model (c). The attention model is used to pick the obstacle, and then the transport one will find the location to place it. This is done by using using the Q function to find the place where the maximum of the Q value is. As a result, the obstacle will be placed at the right place in (d). }
\label{fig:processingheatmap}
\end{figure}

\begin{figure}
\centering
\includegraphics[width=0.45\textwidth]{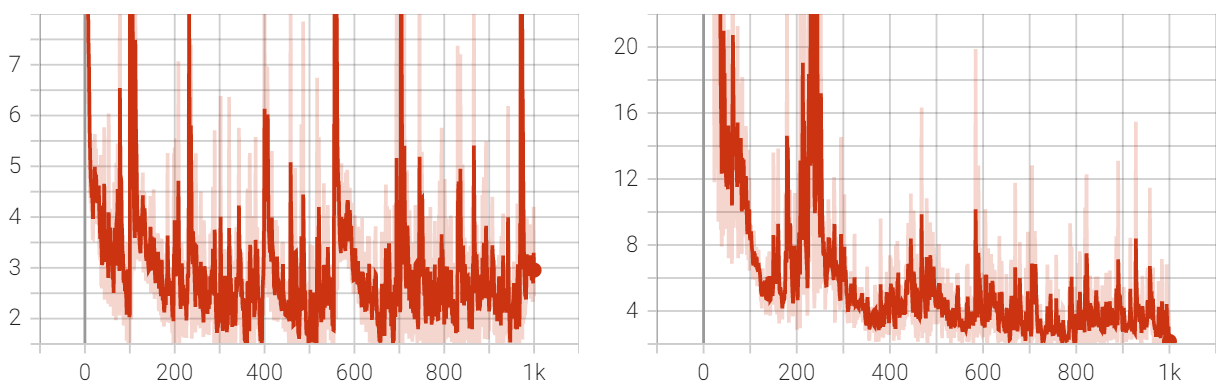}
\caption{Training curves of the picking model (left) and the placing model (right). The loss value of the attention model is relatively small but fluctuates during the training process, while the loss value for the transport model drops significantly. }
\label{fig:loss}
\end{figure}

\textbf{Loss} Training curves of the picking model (attention model) and the placing model (transport model) are shown in Fig. \ref{fig:loss}. The loss value of the attention model fluctuates during the training process, while the loss value for the transport model drops significantly. The comparison between these two curves reveals the different levels of difficulty for the picking and placing tasks and the complexity of the two models.

\textbf{The reward}  We have evaluated their performance based on the average reward gained during the 100 times test. The reward refers to how far the length of the path from anode to cathode, and the range is 0-1. The closer the final path length is to the direct length from the anode to the cathode, the greater the reward will be.

\textbf{Performance Evaluation} Overall, we can see the performance improves, as the reward is increasing when we increase the number of demonstrations trained. The more difficult the task is, the lower reward they will perform if there are only a few demonstrations for training. However, they all perform relatively well as the rewards are close to 1 when the network has been trained with 100 demonstrations.

Among these results, remove-obstacle, which is the simplest task, achieves the best reward all the time, which means it performs perfectly.

\begin{table*}
\label{simulatiom_result_table}
\caption{The network is trained with 1, 10, and 100 demos respectively for these tasks. After that, we evaluate their performance based on the average reward gained during 100 tests. The range of reward is from 0-1}
\begin{tabular*}{\textwidth}{@{\extracolsep{\fill}}ccccccc}
\toprule
  & & forbidden& inverted& \multicolumn{3}{c}{average reward with n demos trained}  \\ \cmidrule(r){5-7}
  Task& obstacles&  zones& electrodes&n = 1 &n = 10&n = 100 \\
\midrule
remove-obstacle& 1& 0& \textcolor{red}{\xmark}& 1.00& 1.00& 1.00 \\
remove-obstacles& 3& 0& \textcolor{red}{\xmark}& 0.94& 0.79& 0.97 \\
bridge-forbidden-zone & 1& 0& \textcolor{red}{\xmark}& 0.98& 0.98& 0.94 \\
bridge-forbidden-zones & 1& 0& \textcolor{red}{\xmark}& 0.86& 0.90& 0.95 \\
draw-above-circuit& 0& 0& \textcolor{green}{\checkmark}& 0.71& 0.87& 0.92 \\
remove-obstacle\\
+ bridge-forbidden-zone& 2& 1& \textcolor{red}{\xmark}& 0.74& 0.81& 0.88 \\
all-in-one& 2& 1& \textcolor{green}{\checkmark}& 0.30& 0.42& 0.77\\
\bottomrule

\end{tabular*}
\end{table*}

\subsection{Real-World Experiments}
\begin{figure*}[htb]
\centerline{\includegraphics[width=1\textwidth]{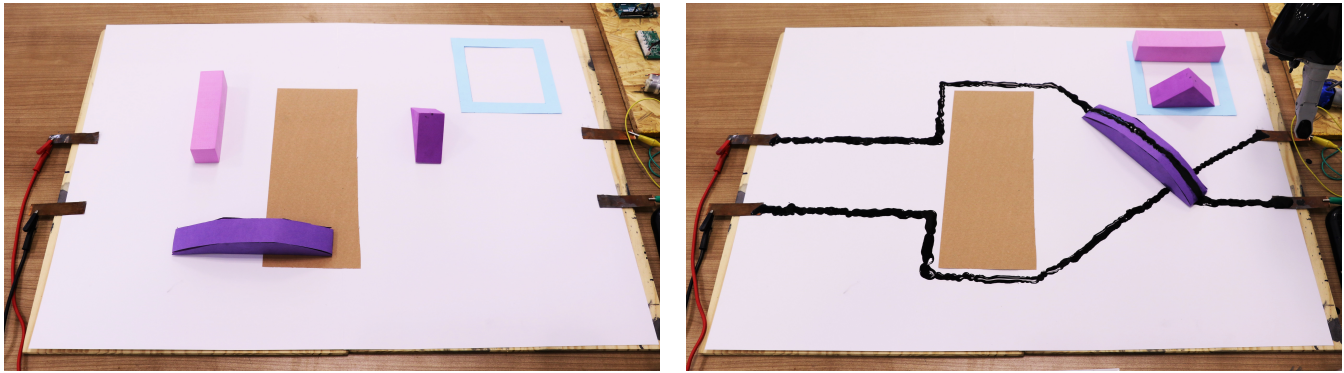}}
\caption{Demonstrations of real-world performance. The left figure shows the initial setting of the workspace. Two obstacles (a triangular prism and a cuboid) and a ramp are placed in the middle area of the cardboard. A forbidden zone (brown rectangle) is set that no circuit can be drawn directly on it. The position and size of all elements are set exactly as same as the setting in the successful simulation test, in which the obstacles are removed to the waste area (blue square) and the ramp is placed in front of the robot terminals to allow circuit crossing on top of each other. The right figure shows the result of real-world actions, where the robot achieves similar results as in the simulation. The battery connected to the metal bars near the robot side starts to recharge after around 30 minutes, showing a good connection of the drawn circuit.}
\label{fig:real_drawing}
\end{figure*}

\begin{figure}[t!]
\centerline{\includegraphics[width=0.48\textwidth]{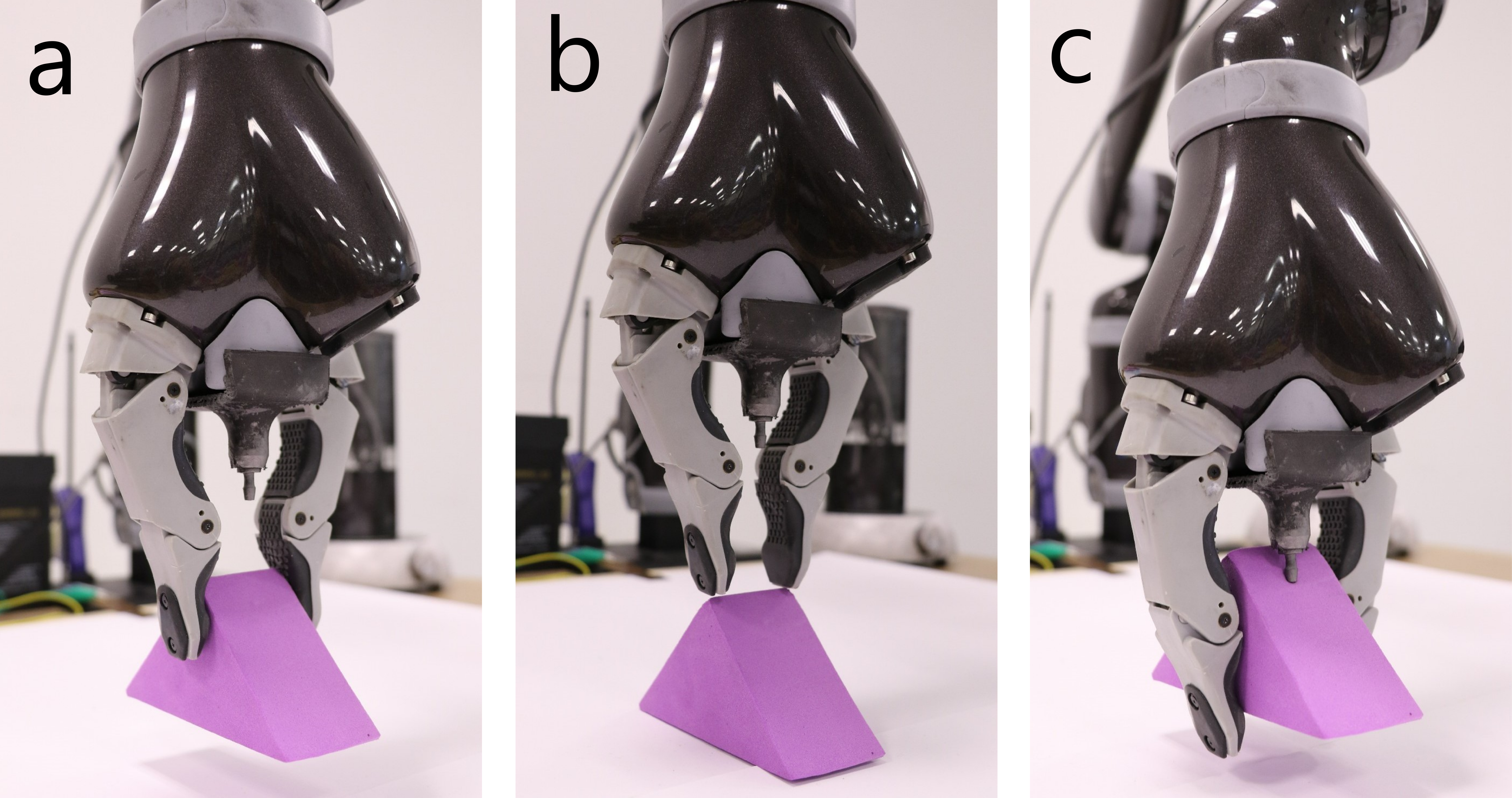}}
\caption{Examples of pick actions. Successful (a), Failed - gripper position too high, can not pick objects (b), Failed - gripper position too low, objects will bend and damage the nozzle (c).}
\label{fig:pick_cases}
\end{figure}

The real-world demonstration is shown in Fig.~\ref{fig:real_drawing}, although the final placement of the objects and circuit are the same as the simulation result, we confronted difficulties when the robot picked objects that we had to repeat the pick action several times. Examples of successful and failed cases are shown in Fig.~\ref{fig:pick_cases}. Overall, the pick action achieves around 60\% success rate from 20 trials. 

\section{DISCUSSION}

\subsection{Simulation Gap in Real-World Experiments}
The main issue in real-world implementation is the misplacement in z-axis of the gripper. The original z value is generated from the network and worked well in simulation tasks. In simulation a constrain between the object and the gripper is automatically created when the object is within a small distance threshold from the gripper to guarantee a firm grasp. However, in real cases, tolerance for displacement in z-axis is much smaller than in the simulation. In the original paper of Transporter \cite{zeng2020transporter}, authors use a suction gripper instead of our two-finger gripper, which is much easier to control but has the limitation that the object surface should be regular. Although the success rate for picking in our real-world experiments is around 60\%, the use of a finger gripper is necessary considering irregular objects and structures in real-world scenarios. 

\subsection{Importance of Circuit Drawing for Future Robots}
To the best of our knowledge, this is the only work where a robot learns to rearrange visual world to receive energy from self-printed circuits. In previous approaches with robots charging themselves, works like \cite{5509556, 5509643} demonstrated the ability of robots plugging themselves to recharge. However, robots can not always rely on outlets to recharge themselves (e.g. search and rescue missions and other unstructured environments). The circuits drawn with conductive ink have much higher flexibility and can be three-dimensionally printed on walls and ceilings, further increasing the likelihood of robots connecting themselves to potential power sources and staying functional. In other works on harvesting energy with specially designed hardware \cite{Xu2010, doi:10.1021/acsenergylett.9b02661}, the accumulation of energy is either slow or low in quantity. Meanwhile, the learning aspect of Robotics within their approaches was left aside. The crucial difference between those and the current work is that in our work the robot needs to learn to rearrange and interact with the environment to access power, while in the previous works this feature is already embedded to those robots.

\subsection{Study Limitations}
In our experiment, the robotic arm has a limited workspace at which the terminal of the power supply is accessible at all times. This condition may differ from a real-case where this robot needs to power itself. Mobile robots equipped with a drawing system, instead of a liquid ink, could be a better fit for real-cases. Another limitation is that the robot is not directly powered by the drawn circuit , and this is because the graphene-based conductive ink we used has a lower conductivity than metal wires (when our robot moves many joints at the same time the current requirement exceeds the maximum current). As material science advances, a conductive ink with a higher conductivity could directly power robots as an alternative to wires or cables, providing a creative way for robots to access ambient energy and stay functional.

\section{CONCLUSION}
This paper presents the first robot capable of accessing energy intake through self-drawn conductive-ink-based electrical connections. With a small number of demonstrations, the robot learns to rearrange surroundings by picking and placing objects to create circuit paths with a minimum amount of conductive ink. Although our experiments are performed in a simplified environment, using conductive ink (a mean) and a power source (an end), this cable-free approach enables flexible deployment in complex environments, which is a very useful proof of concept for autonomous robots (e.g. in search-and-rescue missions). As robots become more present in our society and even reach other planets, maximizing the capacity to stay alive is crucial to increase the odds of success.









\addtolength{\textheight}{-8.5cm}

\bibliographystyle{ieeetr}
\bibliography{root}

\end{document}